# Task Complexity Matters: An Empirical Study of Reasoning in LLMs for Sentiment Analysis

Donghao HUANG[1,2][0009−0005−6767−4872] and
Zhaoxia WANG[1][0000−0001−7674−5488]*

[1] School of Computing and Information Systems, Singapore Management University,
80 Stamford Rd, Singapore 178902, Singapore
`dh.huang.2023@engd.smu.edu.sg, zxwang@smu.edu.sg`
[2] Research and Development, Mastercard, 4250 Fairfax Dr, Arlington, VA 22203,
USA
`donghao.huang@mastercard.com`

**Abstract.** Large language models (LLMs) with reasoning capabilities have fueled a compelling narrative that reasoning universally improves performance across language tasks. We test this claim through a comprehensive evaluation of 504 configurations across seven model families—including adaptive, conditional, and reinforcement learning-based reasoning architectures—on sentiment analysis datasets of varying granularity (binary, five-class, and 27-class emotion). Our findings reveal that reasoning effectiveness is strongly task-dependent, challenging prevailing assumptions: (1) Reasoning shows task-complexity dependence—binary classification degrades up to $-19.9$ F1 percentage points (pp), while 27-class emotion recognition gains up to $+16.0$ pp; (2) Distilled reasoning variants underperform base models by 3–18 pp on simpler tasks, though few-shot prompting enables partial recovery; (3) Few-shot learning improves over zero-shot in most cases regardless of model type, with gains varying by architecture and task complexity; (4) Pareto frontier analysis shows base models dominate efficiency-performance trade-offs, with reasoning justified only for complex emotion recognition despite $2.1\times-54\times$ computational overhead. We complement these quantitative findings with qualitative error analysis revealing that reasoning degrades simpler tasks through systematic over-deliberation, offering mechanistic insight beyond the high-level overthinking hypothesis.

## Introduction

The emergence of reasoning-enhanced large language models has fundamentally transformed our understanding of artificial intelligence capabilities, with models like OpenAI's O1, DeepSeek-R1, and Mistral's Magistral achieving human-level performance on complex cognitive tasks [1].

This success has contributed to a compelling narrative: that enhanced reasoning improves model performance across a wide range of natural language tasks, aligning with the broader goal of developing reasoning-capable LLMs. However, this narrative remains largely untested for foundational NLP applications such as sentiment analysis.

---

* Corresponding author: zxwang@smu.edu.sg

Recent work provides early evidence of this mismatch: Li et al. [6] demonstrate that chain-of-thought reasoning leads to "overthinking" and degraded performance in financial sentiment analysis. This suggests a task-complexity mismatch—reasoning designed for complex problem-solving may introduce unnecessary overhead for tasks that benefit from direct pattern recognition.

Building on the initial investigation by Huang & Wang [4], we present the first large-scale, cross-architectural comparison—including adaptive (Qwen3), conditional (Granite), and RL-based (Magistral) models—and directly compare reasoning-augmented models with their base counterparts to isolate the impact of reasoning enhancements. Our central question: Do reasoning capabilities justify their computational overhead for sentiment tasks of varying complexity?

Our study makes several key contributions:

- We present the first comprehensive analysis comparing thinking/reasoning and non-thinking/base models across sentiment tasks, showing that reasoning often degrades performance—particularly in simple binary classification.
- We show that distilled reasoning variants underperform their base models, especially on simple sentiment analysis tasks, which contrasts with the widely held belief that advanced reasoning capabilities improve LLM performance.
- We demonstrate that few-shot learning yields more robust and consistent improvements than reasoning modes, positioning it as a superior strategy for enhancing model performance in sentiment analysis.
- We conduct the first ever Pareto frontier analysis of performance vs. computational cost for reasoning vs. non-reasoning models in sentiment analysis tasks, providing an objective framework for evaluating cost-performance trade-offs.
- We provide qualitative error analysis explaining *why* reasoning degrades performance on simpler tasks through systematic over-deliberation, offering mechanistic insight that complements our quantitative findings.

To support future research, we release all datasets, prompt templates, code, and complete performance results at GitHub[3].

## Related Work

**Reasoning** has proven effective for mathematical and logical tasks [13], but effectiveness varies with task structure and complexity [8]. Most relevant to our work, Li et al. [6] demonstrate that CoT impairs financial sentiment analysis through "overthinking"—generating elaborate reasoning chains that introduce errors rather than clarifying decisions. Their findings suggest a task-complexity mismatch: reasoning mechanisms optimized for multi-step problem decomposition may be counterproductive for classification tasks where correct answers emerge from direct pattern recognition.

---
[3] Public Repo: https://github.com/inflaton/Task-Complexity-Matters

**Reasoning distillation** efforts explore compressing reasoning into smaller models. Xu et al. [14] find preserving reasoning structure is critical, with structural corruption degrading performance by 12–14%. The DeepSeek-R1 series maintains reasoning with fewer parameters [1], but task alignment remains crucial. Zhang et al. [16] achieve strong results through targeted sentiment distillation, while Hsieh et al. [3] distill mathematical reasoning effectively, though such gains may not transfer to affective tasks.

**Thinking/reasoning architectural innovations** manage reasoning complexity through adaptive compute. Qwen3 introduces "thinking budgets" that adjust dynamically based on query difficulty [9], Granite3.3 separates intermediate reasoning from final answers [5], and Magistral uses reinforcement learning to shape reasoning paths [10].

**Sentiment analysis with LLMs** has attracted increasing attention from researchers [12]. Zhang et al. [15] show that LLMs excel at binary sentiment classification (92%+) but struggle with fine-grained emotion detection. Huang and Wang [4] report strong binary performance for DeepSeek-R1 (99.31%) but note scalability issues and omit base-model comparisons. Building on this early work, we provide the first large-scale, cross-architectural study—including adaptive (Qwen3), conditional (Granite), and RL-based (Magistral) models—and directly compare reasoning-augmented models with their base variants to isolate the effects of reasoning.

## Methodology

**Models.** We evaluate seven model families:

(1) **DeepSeek-R1 (DSR1) series**: full model (671B) and distilled variants (8B, 14B, 32B, 70B);
(2) **DeepSeek-V3 (DSV3)**: base model for DSR1-Full;
(3) **LLaMA**: 3.1–8B and 3.3–70B, serving as bases for DSR1-8B and DSR1-70B;
(4) **Qwen2.5**: 14B and 32B, serving as bases for DSR1-14B and DSR1-32B;
(5) **Qwen3**: dense models (4B, 8B, 14B, 32B), each supporting thinking (T) and non-thinking (N) modes;
(6) **Granite3.3**: lightweight models (2B, 8B) with conditional reasoning in T and N modes;
(7) **Magistral**: 24B model with RL-based reasoning in T and N modes.

For DSR1 full and distilled models, we compare performance directly with their corresponding base models (groups 1–3). For the remaining families, we assess reasoning effects by contrasting their thinking (T) and non-thinking (N) modes. We analyze these architectures separately, as they embody distinct reasoning paradigms—*built-in reasoning* vs. *runtime-activated reasoning*.

**Datasets.** We evaluate on three benchmarks spanning complexity levels [4]:

- **IMDB Movie Reviews** [11]: Binary sentiment (positive/negative), representing simple classification
- **Amazon Reviews** [7]: Five-class sentiment (strongly negative to strongly positive), representing moderate complexity
- **GoEmotions** [2]: 27 emotion categories, representing high complexity. Since GoEmotions is originally a multi-label dataset, we use only a subset of samples that have a single label.

We selected these three datasets to form a granularity spectrum, with the number of target classes (2, 5, 27) serving as a proxy for task complexity. While the datasets also differ in domain (movies, products, social media) and text characteristics, the systematic gradient in results across complexity levels—consistent across all seven model families—suggests that classification granularity is the primary driver of the observed patterns. The monotonic relationship between class count and reasoning effectiveness across diverse architectures would be unlikely if domain differences were the dominant confound.

**Experimental Settings and Configurations.**

Models assessed under 0–5–10–20–30–40–50-shot settings. Few-shot exemplars randomly sampled (seed=42) from training sets, stratified by class to ensure balanced representation. For thinking-enabled models, both T and N modes evaluated to isolate reasoning effects.

**Prompt Design.** All models received identical structured prompts to ensure comparability. The system prompt template consisted of:

```
You are an advanced sentiment analysis assistant. Your task is to analyze text and provide
  a sentiment rating along with a brief explanation. The sentiment rating should be based
  on a \{scale\}-point scale: \{sentiments\}. Always respond with a JSON object
  containing the sentiment and the explanation.
```

Scale parameters were set according to dataset complexity:

- **IMDB** (binary, scale=2): Sentiments = "Positive or Negative"
- **Amazon** (5-class, scale=5): Sentiments = "Strongly Positive, Positive, Neutral, Negative, or Strongly Negative"
- **GoEmotions** (27-class, scale=27): Sentiments = [27 emotion labels]

For few-shot settings, example input-output pairs were appended to the system prompt. Each example followed this format:

```
- Input: [example text]
- Output: {"sentiment": "[label]", "explanation": ""}
```

The user prompt format for test instances was:

```
- Input: [input text]
- Output:
```

**Infrastructure.** DeepSeek-R1 (full) and DeepSeek-V3 accessed via official DeepSeek APIs. All other models run locally using Ollama v0.6.8 on NVIDIA H100 GPU (96GB VRAM, Ubuntu 24.04.1 LTS). All models evaluated using recommended default hyperparameters per official model documentation.

**Metrics and Evaluation.** We report performance using the F1-score: standard (binary) F1 for IMDB (2 classes), weighted F1 for Amazon (5 classes) and GoEmotions (27 classes). Efficiency is measured via mean per-sample latency (in seconds). Overall, the evaluation spans 504 configurations (7 model families × 3 datasets × 7 shot levels × thinking / non-thinking variants, where applicable).

## Results

### Distilled vs. Base Model Performance

Table 1 compares distilled reasoning models with their base counterparts, revealing systematic performance patterns.

**Zero-shot degradation.** Distilled models underperform base models in zero-shot settings (Table 1), with the largest drops on IMDB (−19.9 pp, DSR1 vs. DSV3) and Amazon (−18.4 pp, DSR1-70B vs. Llama3.3-70B). Mean zero-shot degradation decreases with task complexity: −5.9 pp on IMDB, −7.6 on Amazon, −2.0 on GoEmotions.

**Few-shot recovery.** Gaps narrow with examples: IMDB from −5.9 to −1.3 pp; Amazon from −7.6 to −5.2 pp. GoEmotions shows reversals: DSR1-32B (+3.2), DSR1-70B (+4.8), and DSR1 (+3.8) outperform bases in few-shot settings, indicating distilled models retain capabilities reactivated through examples on complex tasks. **Reasoning failure rates and task-dependence.** IMDB shows 100% failure rate (10 of 10 comparisons favor base models), Amazon shows 80% (8 of 10), while GoEmotions shows 50% (5 of 10), as shown in Table 1. The decreasing failure rates with task complexity suggest reasoning-focused distillation becomes more beneficial as classification granularity increases, approaching parity on complex 27-class emotion recognition.

**Computational overhead.** As shown in Table 1, speed ratios range 1.1×–20.9×, with highest costs on IMDB (DSR1 vs. DSV3: 20.9×) and Amazon (DSR1-8B: 14.1× at best-shot). The extreme overhead on simple binary tasks contrasts with more reasonable costs on complex emotion recognition (4.3×–6.2× for larger models), further validating that reasoning mechanisms are most effective when task complexity justifies their computational cost.

### Thinking vs. Non-thinking Mode Analysis

Table 2 compares thinking and non-thinking modes across architectures, isolating the effect of dynamic reasoning activation (separate from distillation effects analyzed previously).

**Task-complexity dependence.** Across 42 zero-shot comparisons, thinking helps GoEmotions (12 of 14 positive, 86%), shows mixed results on Amazon (6 of

**Table 1.** Reasoning-Focused Distillation vs. Base Model Performance. Zero-shot (**0S**) and best-shot (**BS**) F1 (%) comparison. **Diff.** = F1_Dist. − F1_Base. **SR**: mean per-sample latency ratio (values > 1 mean distilled models are slower). **FR**: percentage of cases where distilled models underperform base models (Diff. ≤ 0).

| Dataset | Model | Base | | Dist. | | Diff. | | SR | | FR |
|---|---|---|---|---|---|---|---|---|---|---|
| | | 0S | BS | 0S | BS | 0S | BS | 0S | BS | (%) |
| **IMDB** | Llama3.1-8B vs DSR1-8B | **97.6** | 97.8 | 92.2 | 95.7 | -5.3 | -2.1 | 2.4 | 6.1 | **100** (10/10) |
| | Qwen2.5-14B vs DSR1-14B | 97.3 | **99.1** | 95.8 | 97.5 | -1.5 | -1.6 | 3.0 | 4.2 | |
| | Qwen2.5-32B vs DSR1-32B | 97.8 | 98.4 | 96.0 | 97.1 | -1.8 | -1.3 | 2.8 | 3.6 | |
| | Llama3.3-70B vs DSR1-70B | 98.0 | **99.3** | 97.1 | 98.0 | **-0.9** | -1.3 | 3.5 | 4.7 | |
| | DSV3 vs DSR1 | 97.9 | **99.3** | 78.0 | **99.3** | -19.9 | **-0.0** | 20.9 | 5.8 | |
| **Amazon** | Llama3.1-8B vs DSR1-8B | 67.2 | 75.8 | 58.3 | 69.6 | -8.9 | -6.2 | 5.0 | 14.1 | **80** (8/10) |
| | Qwen2.5-14B vs DSR1-14B | 74.7 | **86.5** | 61.9 | 78.5 | -12.8 | -8.0 | 6.5 | 8.0 | |
| | Qwen2.5-32B vs DSR1-32B | 63.8 | 84.3 | 69.0 | 80.5 | +**5.2** | -3.9 | 4.9 | 5.2 | |
| | Llama3.3-70B vs DSR1-70B | **82.5** | **86.5** | 64.1 | 73.8 | -18.4 | -12.7 | 5.3 | 5.9 | |
| | DSV3 vs DSR1 | 81.6 | 86.6 | 78.3 | **91.4** | -3.3 | +**4.8** | 1.1 | 1.4 | |
| **Go-Emotions** | Llama3.1-8B vs DSR1-8B | 27.9 | 33.4 | 21.0 | 32.2 | -6.9 | -1.1 | 8.1 | 11.5 | **50** (5/10) |
| | Qwen2.5-14B vs DSR1-14B | 38.0 | 39.6 | 34.0 | 38.7 | -3.9 | -0.9 | 4.8 | 4.8 | |
| | Qwen2.5-32B vs DSR1-32B | 34.0 | 37.9 | 34.7 | **41.1** | +0.7 | +3.2 | 4.8 | 4.2 | |
| | Llama3.3-70B vs DSR1-70B | 38.6 | 39.8 | 34.0 | **44.6** | -4.6 | +**4.8** | 4.9 | 6.2 | |
| | DSV3 vs DSR1 | 39.1 | **42.2** | **42.8** | **46.0** | +**3.7** | +3.8 | 5.1 | 4.3 | |

Title Suppressed Due to Excessive Length    7

**Table 2.** Thinking vs. Non-thinking Mode Performance. Zero-shot (**0S**) and best-shot (**BS**) F1 (%) comparison. **Diff.** = F1_Thinking − F1_Non-thinking. **SR**: latency ratio (Thinking/Non-thinking; values > 1 mean thinking is slower). **FR**: percentage of cases where thinking models underperform non-thinking models (Diff. ≤ 0).

| Dataset | Model | Non-think. 0S | Non-think. BS | Thinking 0S | Thinking BS | Diff. 0S | Diff. BS | SR 0S | SR BS | FR (%) |
|---|---|---|---|---|---|---|---|---|---|---|
| IMDB | Granite 3.3-2B | 95.3 | 95.3 | 95.4 | 95.4 | +0.1 | +0.1 | 3.8 | 3.8 | **36** (5/14) |
| | Qwen3-4B | 97.0 | 98.2 | 97.5 | 98.4 | +0.5 | +0.2 | 4.3 | 5.5 | |
| | Granite 3.3-8B | 96.6 | 98.3 | 95.7 | 96.0 | -0.9 | -2.2 | 3.2 | 1.0 | |
| | Qwen3-8B | 98.2 | 98.6 | 97.7 | 98.2 | -0.5 | -0.5 | 2.5 | 1.2 | |
| | Qwen3-14B | 96.5 | 98.4 | 98.2 | 98.9 | +1.6 | +0.5 | 3.9 | 4.3 | |
| | Magistral-24B | 94.8 | 97.4 | 96.9 | 96.9 | +2.1 | -0.5 | 12.2 | 9.7 | |
| | Qwen3-32B | 98.2 | 98.8 | 98.6 | 98.9 | +0.5 | +0.0 | 3.9 | 4.3 | |
| Amazon | Granite 3.3-2B | 74.1 | 74.1 | 63.9 | 63.9 | -10.2 | -10.2 | 2.1 | 2.1 | **57** (8/14) |
| | Qwen3-4B | 75.9 | 83.8 | 69.9 | 84.3 | -6.0 | +0.4 | 7.7 | 8.3 | |
| | Granite 3.3-8B | 79.8 | 82.4 | 77.3 | 78.0 | -2.5 | -4.3 | 1.8 | 2.4 | |
| | Qwen3-8B | 71.1 | 84.5 | 77.9 | 84.8 | +6.8 | +0.3 | 8.1 | 5.4 | |
| | Qwen3-14B | 70.3 | 84.1 | 74.5 | 84.6 | +4.2 | +0.5 | 4.4 | 6.1 | |
| | Magistral-24B | 66.1 | 81.6 | 78.3 | 80.7 | +12.2 | -0.9 | 20.0 | 14.4 | |
| | Qwen3-32B | 75.7 | 89.0 | 74.9 | 84.8 | -0.7 | -6.8 | 2.4 | 11.5 | |
| Go-Emotions | Granite 3.3-2B | 32.2 | 32.2 | 26.1 | 28.6 | -6.1 | -3.6 | 5.3 | 3.2 | **14** (2/14) |
| | Qwen3-4B | 37.5 | 42.8 | 38.7 | 43.4 | +1.1 | +0.6 | 6.5 | 9.2 | |
| | Granite 3.3-8B | 28.9 | 37.3 | 31.6 | 39.4 | +2.8 | +2.2 | 5.5 | 1.4 | |
| | Qwen3-8B | 35.8 | 41.0 | 38.4 | 44.0 | +2.6 | +3.1 | 7.6 | 8.3 | |
| | Qwen3-14B | 36.1 | 40.3 | 42.6 | 47.0 | +6.5 | +6.6 | 8.0 | 8.4 | |
| | Magistral-24B | 22.0 | 35.0 | 38.0 | 43.7 | +16.0 | +8.8 | 25.7 | 54.4 | |
| | Qwen3-32B | 34.7 | 39.8 | 41.6 | 45.9 | +6.9 | +6.2 | 6.2 | 7.4 | |

**Table 3.** Aggregated Performance Across 504 Configurations. F1 (%) mean ± std across 7 shot levels and 12 model pairs (5 from Table 1, 7 from Table 2). **N**: number of configurations. **Diff.** = F1_Dist./Think. − F1_Base/Non-think.

| Dataset | Type | N | F1 (%, mean±std) | Diff. (%, mean±std) |
|---|---|---|---|---|
| **IMDB** | Base/Non-think. | 84 | $91.5 \pm 20.8$ | $-4.8 \pm 6.3$ |
| | Dist./Think. | 84 | $86.7 \pm 25.7$ | |
| **Amazon** | Base/Non-think. | 84 | $74.9 \pm 18.9$ | $-3.6 \pm 2.2$ |
| | Dist./Think. | 84 | $71.3 \pm 17.7$ | |
| **GoEmotions** | Base/Non-think. | 84 | $36.6 \pm 4.0$ | $+2.0 \pm 1.0$ |
| | Dist./Think. | 84 | $38.6 \pm 7.3$ | |

14 positive, 43%), and balanced effects on IMDB (9 of 14 positive, 64%). Failure rates by dataset: Amazon 57% (8 of 14), IMDB 36% (5 of 14), GoEmotions 14% (2 of 14). The substantially lower failure rate for complex emotion recognition (14%) compared to simpler tasks (36-57%) confirms that reasoning mechanisms are most effective when task complexity justifies their computational cost.

**Few-shot convergence varies by complexity.** Thinking advantages often diminish with examples: Magistral shifts from $+12.2$ pp to $-0.9$ pp on Amazon, and $+2.1$ pp to $-0.5$ pp on IMDB (Table 2). GoEmotions maintains consistent thinking advantages (mean $+4.3$ pp zero-shot to $+3.4$ pp best-shot), while IMDB and Amazon show thinking models losing ground in few-shot settings. This suggests simpler tasks benefit more from direct pattern matching with exemplars than from reasoning overhead.

**Efficiency trade-offs.** Mean speed ratios vary by architecture: Granite 1.0×–5.5×, Qwen3 2.5×–11.5×, Magistral 9.7×–54.4× (Table 2). GoEmotions consistently shows highest overhead, indicating reasoning cost scales non-linearly with task complexity.

### Aggregate Performance by Task Complexity

Table 3 presents aggregated metrics across all 504 configurations, comparing base/non-thinking models against distilled/thinking variants.

**Task-dependent reasoning effectiveness.** Aggregating across all 504 configurations reveals systematic task-complexity patterns. As shown in Table 3, base/non-thinking models outperform distilled/thinking variants on simpler tasks: IMDB shows $4.8 \pm 6.3$ percentage points (pp) advantage, Amazon shows $3.6 \pm 2.2$ pp advantage. This pattern reverses for complex emotion recognition, where distilled/thinking models achieve $2.0 \pm 1.0$ pp advantage on GoEmotions. The decreasing performance gap with increasing task complexity—from $-4.8$ pp (binary) through $-3.6$ pp (5-class) to $+2.0$ pp (27-class)—validates the hypothesis that reasoning mechanisms become more beneficial as classification granularity increases.

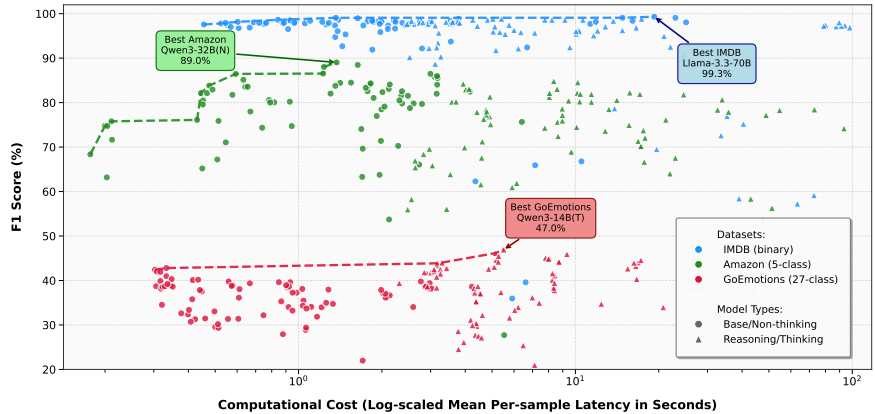

**Fig. 1.** Performance vs. computational cost. Circles: base/non-thinking; triangles: reasoning/thinking. Colors: IMDB (blue), Amazon (green), GoEmotions (red). Dashed lines: Pareto frontiers. Evaluated on H100 via Ollama; DSR1/V3 excluded.

**Performance variability and model selection.** As shown in Table 3, reasoning approaches introduce greater performance variance: on IMDB, distilled/thinking models show higher variability (SD: 25.7 pp) than base models (SD: 20.8 pp), while GoEmotions similarly shows higher variance for reasoning models (7.3 pp vs. 4.0 pp)—though this variability accompanies overall performance gains on complex tasks. Mean F1 scores confirm a clear task-difficulty hierarchy: IMDB achieves the highest performance (91.5% base, 86.7% reasoning), Amazon shows moderate levels (74.9% vs. 71.3%), and GoEmotions the lowest (36.6% vs. 38.6%)—with GoEmotions being the only task where reasoning models outperform base models on average.

The inverse relationship between task complexity and base model advantage provides clear deployment guidance: base/non-thinking models offer superior accuracy with lower overhead for binary classification (4.8 pp advantage), the gap narrows for 5-class sentiment (3.6 pp), and reverses for 27-class emotion recognition ($-2.0$ pp), justifying reasoning only for the most complex scenarios. This consistent pattern across 504 configurations provides robust evidence for task-dependent model selection.

### Pareto Frontier Analysis

Figure 1 visualizes efficiency-performance trade-offs across all 504 configurations.

**Pareto frontiers** identify configurations where no alternative improves F1 without increasing latency, or reduces latency without sacrificing F1. IMDB's frontier is dominated by base models (97–99% F1 with minimal latency) and appears flat, indicating diminishing returns from additional compute. GoEmotions includes reasoning models more frequently, with a steeper cost-performance slope where latency increases yield greater accuracy improvements. Amazon shows mixed representation with peak F1≈89% (Qwen3-32B-N).

## Discussion

**Why Reasoning Fails: A Mechanistic Perspective.** Our findings reveal a systematic relationship between task complexity and reasoning effectiveness that goes beyond the high-level "overthinking" explanation [6]. We examined model outputs across complexity levels to identify concrete failure patterns. On binary sentiment tasks, reasoning models frequently generate elaborate deliberation chains that introduce spurious considerations: when classifying clearly positive reviews, thinking models often hedge by identifying minor negative elements (e.g., acknowledging a subplot weakness) before arriving at an overall verdict—a process that occasionally flips the final classification. This over-deliberation is counterproductive because binary sentiment typically requires recognizing dominant sentiment signals rather than weighing competing evidence.

In contrast, for 27-class emotion recognition, the same deliberative process proves beneficial: distinguishing "disappointment" from "sadness" or "annoyance" from "anger" genuinely requires evaluating subtle contextual cues—precisely the nuanced analysis that reasoning mechanisms facilitate. The systematic relationship between failure rate and task complexity ($100\% \to 80\% \to 50\%$ for distilled models; GoEmotions: 14% for thinking modes vs. 36–57% for simpler tasks) corroborates this interpretation: as the number of decision boundaries increases, the computational cost of deliberation becomes increasingly justified by the need for finer discrimination.

**Architecture Specialization.** Different reasoning approaches show distinct profiles: RL-based (Magistral) provides strong zero-shot benefits but loses advantage with examples; adaptive budgets (Qwen3) maintain consistent modest benefits; conditional reasoning (Granite) struggles on simpler tasks but improves on complex ones. This suggests reasoning mechanisms should match task cognitive demands.

**Reasoning Distillation Limitations.** The systematic zero-shot degradation of distilled models indicates reasoning patterns optimized for multi-step problem-solving transfer poorly to sentiment classification. However, few-shot recovery on GoEmotions suggests distilled models retain linguistic capabilities reactivatable through examples on sufficiently complex tasks, consistent with our mechanistic analysis above.

**Practical Deployment.** Pareto analysis shows that base/non-thinking models dominate for most sentiment classification tasks. Few-shot learning (e.g., the best-shot setting) exhibits task-dependent behavior: it provides gains on both complex (emotion recognition) and simpler (binary classification) tasks, irrespective of the model's reasoning mode. In practice, reasoning-capable models should be reserved for complex multi-class emotion tasks, where the accuracy improvements justify their 2.1×–54× computational overhead.

**Limitations.** All experiments use English-language benchmarks; cross-linguistic validation remains open. Our scope is limited to sentiment analysis—extending to tasks like natural language inference would further test generality. GoEmotions is evaluated in a single-label setting, which may underestimate the original multi-label difficulty. We operationalize "reasoning" through distilled models and thinking/non-thinking modes rather than explicit chain-of-thought prompting [13], a distinct paradigm. Finally, alternative prompt designs (e.g., disabling explanation generation) might mitigate degradation on simpler tasks. Despite these limitations, consistency across seven model families and 504 configurations provides robust evidence for task-complexity-dependent reasoning effectiveness.

## Conclusion

We provide the first large-scale, systematic assessment of how reasoning-capable LLMs impact sentiment analysis across varying complexity levels. Our evaluation of 504 configurations demonstrates that reasoning effectiveness is task-dependent: simpler tasks suffer degradation while complex 27-class emotion recognition benefits. Distilled reasoning variants generally underperform base counterparts, though few-shot prompting partially mitigates this gap. Few-shot learning emerges as a consistent strength across both reasoning and non-reasoning models. Our Pareto analysis confirms that base models dominate the efficiency–performance landscape for most sentiment tasks, with reasoning justifiable only for high-complexity scenarios.

These insights call for more nuanced narratives about LLM reasoning. Future work should explore: (1) hybrid architectures that trigger reasoning only when ambiguity warrants it; (2) task-specific distillation for affective reasoning; (3) cross-linguistic validation; (4) extension to other NLP classification tasks to test generality; and (5) alternative prompt designs that may mitigate reasoning overhead on simpler tasks.

12      D. Huang et al.